\documentclass[10pt,twocolumn,letterpaper]{article}

\usepackage{iccv}
\usepackage{times}
\usepackage{epsfig}
\usepackage{graphicx}
\usepackage{amsmath}
\usepackage{amssymb}
\usepackage{float}
\usepackage{stfloats}
\usepackage[ruled,vlined]{algorithm2e}

\newcommand{\supp}[1]{\textit{\textcolor{magenta}{#1}}}

\usepackage[pagebackref=true,breaklinks=true,letterpaper=true,colorlinks,bookmarks=false]{hyperref}

\iccvfinalcopy

\ificcvfinal\fi

\begin{document}

\title{Out-of-boundary View Synthesis Towards Full-Frame Video Stabilization}

\author{
Yufei Xu$^1$ \quad
Jing Zhang$^1$ \quad
Dacheng Tao$^{1,2}$\\
$^1$The University of Sydney, Australia\\
$^2$JD Explore Academy, China \\
{\tt\small yuxu7116@uni.sydney.edu.au} \quad
{\tt\small jing.zhang1@sydney.edu.au} \quad
{\tt\small dacheng.tao@gmail.com}
}

\maketitle

\begin{abstract}

Warping-based video stabilizers smooth camera trajectory by constraining each pixel's displacement and warp stabilized frames from unstable ones accordingly. However, since the view outside the boundary is not available during warping, the resulting holes around the boundary of the stabilized frame must be discarded (i.e., cropping) to maintain visual consistency, and thus does leads to a tradeoff between stability and cropping ratio. In this paper, we make a first attempt to address this issue by proposing a new \textbf{O}ut-of-boundary \textbf{V}iew \textbf{S}ynthesis (OVS) method. By the nature of spatial coherence between adjacent frames and within each frame, OVS extrapolates the out-of-boundary view by aligning adjacent frames to each reference one. Technically, it first calculates the optical flow and propagates it to the outer boundary region according to the affinity, and then warps pixels accordingly. OVS can be integrated into existing warping-based stabilizers as a plug-and-play module to significantly improve the cropping ratio of the stabilized results. In addition, stability is improved because the jitter amplification effect caused by cropping and resizing is reduced. Experimental results on the NUS benchmark show that OVS can improve the performance of five representative state-of-the-art methods in terms of objective metrics and subjective visual quality.\footnote{The code is publicly available at \href{https://github.com/Annbless/OVS\_Stabilization}{code}.}
   
\end{abstract}

\section{Introduction}

\begin{figure}[htbp]
    \centering
    \includegraphics[width=1\linewidth]{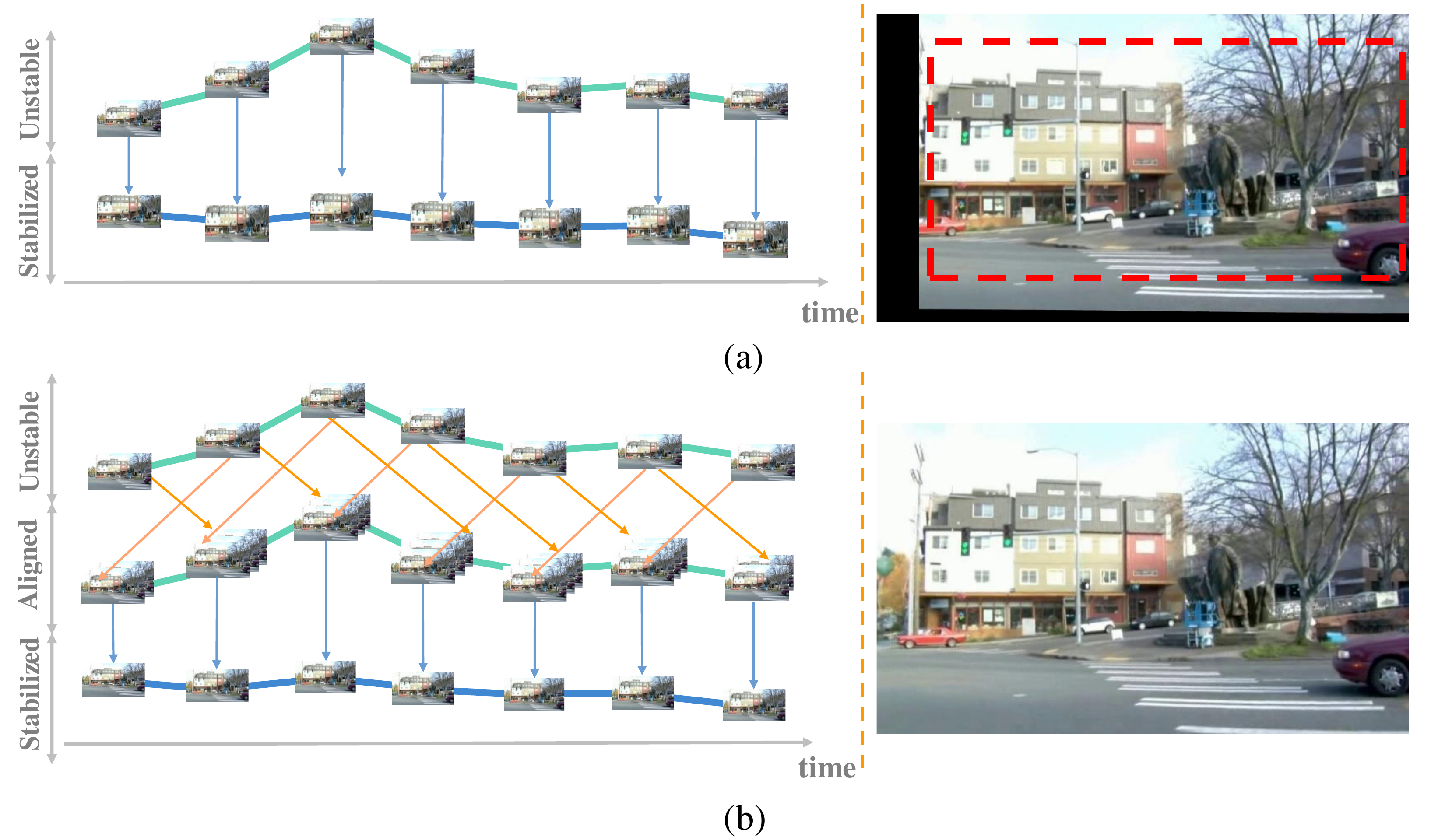}
    \caption{(a) Previous warping-based video stabilization methods only use the current frame to get the stabilized frame by warping. Since the view outside the boundary is not available during warping, the resulting holes around the boundary of the stabilized frame must be cropped. (b) Our OVS method leverages neighboring frames to expand the area of the out-of-boundary view of each current frame. Therefore the subsequent warping process can find requisite candidate pixels for obtaining a full-frame stable frame.}
    \label{fig:opening}
\end{figure}

With the increased demand for high-quality video using handheld devices, video stabilization has become increasingly important, as such videos always contain undesirable jitter. Many video stabilization methods have been proposed to eliminate jitter in unstable videos for a better visual experience~\cite{wang2018deep,Choi_TOG20,yu2020learning,yu2018selfie,yu2019robust}, and can facilitate many other computer vision tasks~\cite{bertinetto2016fully,oh2019video,zhang2020empowering,cheng2020learning,zhang2020towards,zhan2019exploring}.

Warping-based stabilizers~\cite{liu2014steadyflow,liu2016meshflow,wang2018deep,yu2018selfie,xu2018deep,yu2019robust,yu2020learning} perform stabilization by first estimating and then smoothing the camera trajectories. The stabilized video is warped from the unstable video based on the pixel displacement field obtained from the transformation between the shake and smoothed trajectories. Unfortunately, some of the requisite source pixels during warping lie outside the boundary of the current unstable frame, inevitably leading to holes near the boundary of the stabilization result. To maintain visual consistency, cropping and resizing operations are employed to discard these holes, but may result in a reduction of the effective frame size, a change in the frame aspect ratio, and an amplification of the jitter. Previous approaches mitigate this problem by reducing the area of these out-of-boundary pixels, \ie, by limiting the maximum deformation displacement~\cite{liu2009content,liu2011subspace,grundmann2011auto,goldstein2012video}. This constraint makes stability and crop ratio a compromise. It provides smoother trajectories with large cropping area and vice versa, both of which are not ideal for a better visual experience. Is it possible to maintain stability while increasing the cropping ratio to get (near) full frame stabilized results?

The interpolation-based stabilizers~\cite{Choi_TOG20} provide a solution to achieve this goal by iteratively interpolating intermediate frames from adjacent frames, including those pixels that lie in the out-of-boundary view of the current frame. These methods implicitly exploit the property that the content in adjacent frames and within each frame follows spatial coherence, a property that has been widely adopted in SFM~\cite{hartley2000multiple,wei2020deepsfm}, video inpainting~\cite{li2019progressive,kim2019deep,zeng2020learning}, and super-resolution~\cite{li2019fast,lim2017enhanced,Zhang_2018_CVPR}. However, this property has not been explored in previous warping-based methods for filling the out-of-boundary pixels. The most intuitive way to exploit this property is to use interpolation to fill the black holes after warping. Unfortunately, some valuable content may have been permanently discarded because it was not sampled during the warping process when the stable frames were obtained, making this post-warping inpainting method invalid here as shown in ~\cite{Choi_TOG20}. In contrast, we try to investigate pre-warping extrapolation to alleviate this problem, which aims to synthesize the out-of-boundary view of each frame by exploiting the property of spatial coherence, thus facilitating the subsequent warping process to sample enough pixels as needed.

Specifically, we propose a new \textbf{O}ut-of-boundary \textbf{V}iew \textbf{S}ynthesis (OVS) method in this paper, which consists of two stages, in which the view outside the boundaries is inferred from the adjacent frames by aligning them to each reference frame. In the first coarse alignment stage, the adjacent frames are roughly aligned with the reference frame using a grid-based motion estimate. Afterwards, a second fine alignment stage is introduced to handle subtle misalignment and refine results.
It first calculates the optical flow, then predicts the optical flow in the out-of-boundary views via affinity propagation, and finally warps pixels from adjacent frames according to the predicted optical flows. This process is iteratively carried out to gradually align distant neighboring frames to the current frame to expand the area of the out-of-boundary view, so that the subsequent warping process can find needed candidate pixels to obtain a stable frame.
Experimental results on the NUS benchmark show that OVS can be plugged in five representative warping-based methods, significantly improving the cropping ratio of stabilized results. 

In summary, the contribution of this work is threefold:

1) We make a first attempt to improve warping-based video stabilizers towards full-frame stabilization by extrapolating the requisite out-of-boundary view during warping.

2) We propose a two-stage coarse-to-fine method for out-of-boundary view synthesis by exploiting the spatial coherence in the video.

3) Experimental results on publicly available datasets show that the proposed method can serve as a plug-and-play module to significantly improve both grid-based and pixel-based warping methods.

\section{Related Work}

\subsection{Warping-based Video Stabilization} 

A representative solution for video stabilization is to estimate the warping field from unstable frames to stabilized ones. Traditional methods typically follow a three-step procedure, first estimating the trajectory, then smoothing the trajectory, and finally obtaining the stabilized frame from the unstable one based on the warping field. The warping field is generated by computing the transformation between the original trajectory and the smoothed trajectory. These methods can constrain the maximum transformation to reduce the area of the out-of-boundary view needed during the warping process, yielding a high cropping ratio while leading to low stability. For example, Subspace~\cite{liu2011subspace} minimizes the displacement of pixels after smoothing the trajectory, which is fitted by a polynomial curve. A similar strategy is utilized in CPW~\cite{liu2009content,grundmann2011auto}. Bundled~\cite{liu2013bundled}, SteadyFlow~\cite{liu2014steadyflow}, and MeshFlow~\cite{liu2016meshflow} try to minimize the transformation as well as the motion between adjacent frames to reduce the area of the out-of-boundary view. Liu~\etal ~\cite{liu2012video} uses a depth camera for video stabilization and limits the maximum rotation and translation transformations. L1Stabilizer~\cite{grundmann2011auto} limits the range of each element in the warping matrix to reduce the area of the out-of-boundary view. Deep learning-based stabilizers learn to regress the unstable-to-stable warping field for stabilization or stabilized frames directly. By leveraging the ground truth stable frames as supervisory signal \cite{wang2018deep,xu2018deep}, the warping field is implicitly constrained to produce stabilized results. Since some pixels on the stabilized frame are not available due to the absence of pixel correspondence during warping, their losses are not calculated, implying that cropping is still needed to get the desired final result.

These methods only make a tradeoff between cropping ratio and stability as the need of out-of-boundary pixels during warping is reduced rather than satisfied. In contrast, we propose a new method, named OVS, that explicitly extrapolates the requisite out-of-boundary pixels for warping, helping warping-based stabilizers to achieve full-frame video stabilization. It offers a new perceptive on video stabilization and does not require the transformation to be limited to a small range. In addition, OVS can serve as a plug-and-play module to significantly improve both grid-based and pixel-based warping methods.

\begin{figure*}[htbp] 
\centering 
    \includegraphics[width=0.95\linewidth]{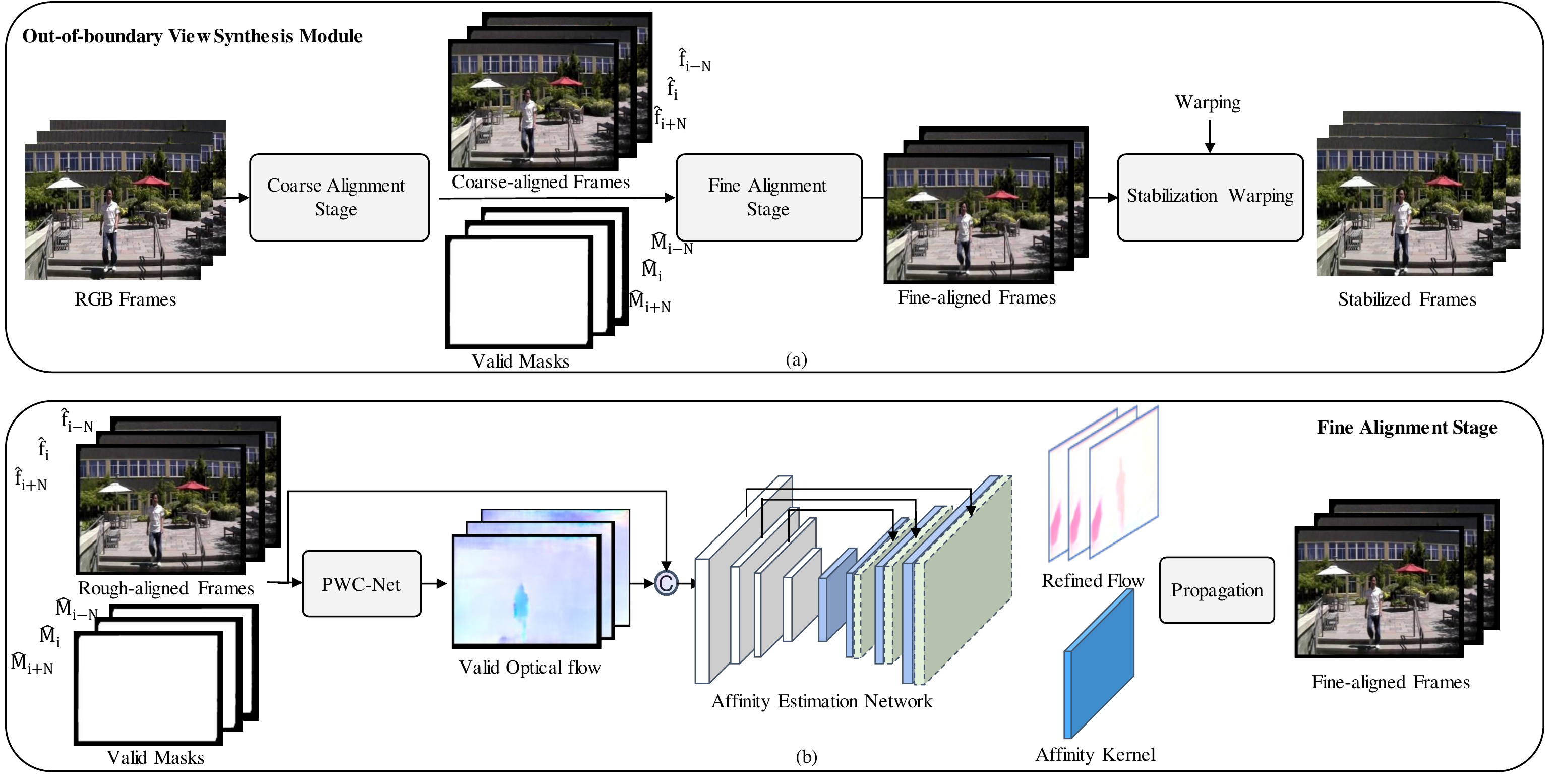}
    \caption{(a) OVS consists of a two alignment stages that align adjacent frames to each reference frame from coarse to fine. (b) The fine alignment stage first estimates optical flow between the roughly aligned frame and the reference frame inside the boundary, then propagates them outside the boundary based on affinity, and finally the out-of-boundary pixels can be extrapolated based on the optical flow. }
    \label{fig:NetworkStructure}
\end{figure*}

\subsection{Interpolation-based Video Stabilization}

Towards cropping-free video stabilization, DIFRINT~\cite{Choi_TOG20} proposes to treat stabilization as a frame generation problem, where the stabilized frames are intermediate frames generated by interpolating from adjacent frames in an iterative manner. However, due to the large jitter in unstable videos, the estimated optical flow for interpolation is not reliable, leading to distortions or ghost artifacts, especially around the boundaries of dynamic objects. Nevertheless, it is insightful to exploit the spatial coherence of adjacent frames for full-frame video stabilization. In this paper, we also investigate the impact of spatial coherence, but from a different perspective, \ie, extrapolating the out-of-boundary view for benefiting warping-based methods rather than interpolation for video stabilization directly. Together with warping-based stabilizers, they can obtain better stabilized results with less distortions or ghost artifacts while maintaining a high cropping ratio as well as good stability.

\subsection{Image Alignment}
Obtaining out-of-boundary view of each frame can be obtained by aligning adjacent frames to the reference one. Traditional image alignment methods always detect feature points first, \eg, SIFT~\cite{lowe2004distinctive}, SURF~\cite{bay2006surf}, ORB~\cite{rublee2011orb}, LIFT~\cite{yi2016lift}, then select robust feature points from them, \eg, by RANSAC~\cite{hartley2000multiple}, and finally use them to calculate the transformation for alignment. Recently, several deep learning-based image alignment methods are proposed~\cite{detone2016deep,nguyen2018unsupervised,zhang2020content} by either supervised learning or unsupervised learning. For example, Zhang \etal~\cite{zhang2020content} estimates a global homography between adjacent frames and uses it for alignment. It is noteworthy that using a single global homography may be ineffective when there is a large movement of the camera pose, which is very common in unstable videos in the video stabilization task. Nevertheless, because of its superiority over traditional methods, we use it as the baseline method for out-of-boundary view synthesis. In contrast, we propose a two-stage coarse-to-fine method, which can deal with large camera movements and dynamic objects more effectively.

\section{Methods}

The proposed OVS method aims to synthesize the out-of-boundary view for video stabilization. It consists of a coarse alignment stage and a fine alignment stage. As shown in Figure~\ref{fig:NetworkStructure}, the coarse alignment stage takes each current frame together with its neighboring frames as input. It aligns each neighboring frame to the reference one and generates a mask to indicate the valid out-of-boundary view after alignment. The second fine alignment stage takes the aligned frames and their masks as input for further refinement. These two stages are carried out alternatively to gradually align distant frames to the reference one and expand the area of out-of-boundary view. The details of the two stages will be discussed as follows.

\subsection{Coarse Alignment}

\begin{figure*}[htbp] 
\centering 
    \includegraphics[width=0.95\linewidth]{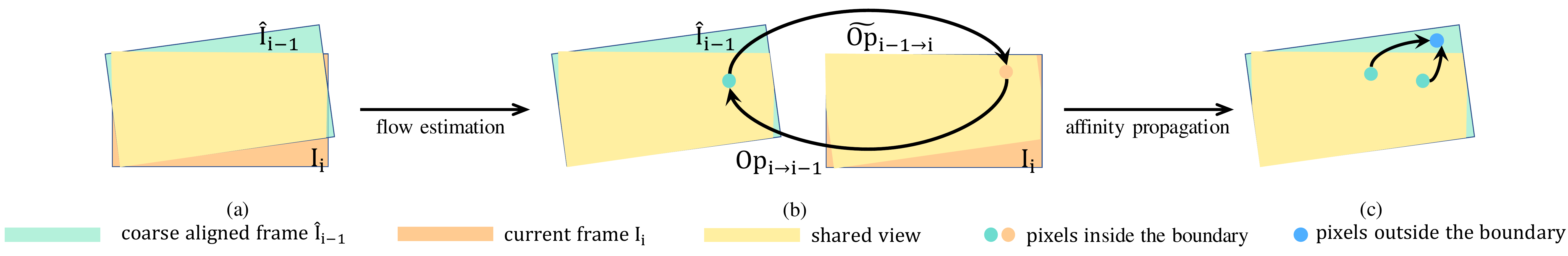}
    \caption{Illustration of the fine alignment stage. (a) The roughly aligned frames from the first stage. (b) Estimation of the optical flow ${Op}_{i \to i-1}$ from the reference frame $I_i$ to the aligned frame $\widehat{I}_{i-1}$ and the reverse flow $\widetilde{Op}_{i-1 \to i}$ from $\widehat{I}_{i-1}$ to $I_i$. (c) The optical flows inside the boundary are propagated outside the boundary using affinity. The out-of-boundary pixels can be extrapolated based on the optical flow.}
    \label{fig:FinePip}
\end{figure*}

We take the motion estimation module in DUT~\cite{xu2020dut} for motion estimation in the coarse alignment stage. It first detects keypoints and estimates their motion in each frame, then propagates the motion of keypoints to a set of predefined grids, and finally estimates the homography for each grid. The keypoints are robust to noise and illumination variation while the grid-based estimation can handle dynamic objects and large jitter. In addition, a multi-planar estimation strategy is used to deal with the complex motion patterns in real scenes. After obtaining the motion of the mesh vertices, the mesh-based homography matrices are computed and used for coarse alignment. The aligned frames are used as the input for the fine alignment stage.  Specifically, the frames in the input unstable video are assumed to be $\{I_i \mid i=1, \cdots, n\}$, where $n$ is the total number of frames and $I_i$ represents the $ith$ frame. The size of the input frames is $H_o \times W_o$, where $H_o$ is 480 and $W_o$ is 640. We employ zero-padding outside the boundary of each frame before warping, \ie 80 pixels in each direction and using masks $\{M_i \mid i=1, \cdots, n\}$ to denote the valid regions after padding. The final output of the coarse alignment module is $\{\hat{I}_{i \to i+1}, \hat{M}_{i \to i+1} \mid i=1, \cdots, n\} \cup \{\hat{I}_{i \to i-1}, \hat{M}_{i \to i-1} \mid i=1, \cdots, n\}$, where the former set is generated from a forward pass and the latter set is generated from a backward pass. Note that we keep $\hat{I}_{n \to n+1} \triangleq I_n$, $\hat{M}_{n \to n+1} \triangleq M_n$, $\hat{I}_{1 \to 0} \triangleq I_1$, and $\hat{M}_{1 \to 0} \triangleq M_1$. The coarse-aligned frames ($\hat{I}$) and corresponding masks ($\hat{M}$) are of size $H_m \times W_m$, where $H_m=640$ and $W_m=800$. 

\subsection{Fine Alignment}

The coarse aligned frames can serve as good initial out-of-boundary view synthesis results, although there is some subtle misalignment due to the influence of dynamic objects or large jitter. To further refine the results, we propose a fine alignment (FA) stage. In the following, we denote $\hat{I}_{i \to i+1}$ as $\hat{I}_i$ and $\hat{M}_{i \to i+1}$ as $\hat{M}_i$ and only present the forward pass for simplicity. As shown in Figure~\ref{fig:NetworkStructure}(b), the fine alignment stage first estimates optical flow between the roughly aligned frame $\widehat{I}_{i-1}$ and the reference frame $I_i$ inside the boundary, then propagates them outside the boundary based on affinity, and finally the out-of-boundary pixels can be extrapolated based on the optical flow from the reference frame as the refined result. 

Taking $\{\widehat{I}_{i-1}, {I_i}\}$ as input, the optical flow ${Op}_{i \to i-1}$ from $I_i$ to $\widehat{I}_{i-1}$ is estimated using PWCNet~\cite{sun2018pwc}:
\begin{equation}
    {Op}_{i \to i-1} = \textit{PWCNet}(I_i, \widehat{I}_{i-1}) \odot M_i,
\end{equation}
where $\odot$ denotes element-wise multiplication. Note that PWCNet has been widely used for optical flow estimation in many stabilizers~\cite{Choi_TOG20,xu2020dut} due to its good performance. Then, a flow reverse (FR) layer~\cite{bao2019depth,jiang2018super} is employed to get the reversed optical flow from $\widehat{I}_{i-1}$ to $I_i$:
\begin{equation}
    (\widetilde{{Op}}_{i-1 \to i}, \widetilde{M}_i) = \textit{FR}({Op}_{i \to i-1}, M_i).
\end{equation}
The reason that we use flow reverse to estimate ${Op}_{{f-1} \to i}$ instead of directly estimating it from $\widehat{I}_{i-1}$ to $I_i$ using PWCNet is that some pixels in $\widehat{I}_{i-1}$ may correspond to the pixels outside the boundary of $I_i$, therefore leading to erroneous optical flow. Moreover, the flow reverse layer can output a mask $\widetilde{M}_i$ which indicates whether or not a pixel in $\widehat{I}_{i-1}$ has a corresponding pixel in $I_i$, \ie, the yellow shared view of $\widehat{I}_{i-1}$ and $I_i$ illustrated in Figure~\ref{fig:FinePip}.

Given the ${Op}_{{i-1} \to i}$ inside the shared view, we want to get the optical flow outside the boundary such that the pixels there can be extrapolated from $I_i$ accordingly. Following the spatial coherence property within the frame, we argue that the motion of static objects should be coherent locally. Based on this assumption, we propose to estimate the affinity kernels from the color and structure information and use them to propagate the optical flow inside the shared view to the outside of the boundary as illustrated in Figure~\ref{fig:FinePip}(c).

Technically, we devise an encoder-decoder network as shown in Figure~\ref{fig:NetworkStructure}(b) to estimate the affinity kernels from $I_i$, $\widehat{I}_{i-1}$, $M_i$, $\widehat{M}_{i-1}$, $\widetilde{M_i}$, $\widetilde{{Op}}_{i-1 \to i}$, $G_i$, and $\widehat{G}_{i-1}$, where $G_i$ ($\widehat{G}_{i-1}$) denotes the edge map extracted from $I_i$ ($\widehat{I}_{i-1}$) using the Sobel filter, which encodes the structure information. We use ResNet-50~\cite{he2016deep} as the backbone encoder. The decoder takes the features from the last layer of the encoder as input and gradually upsamples the features to decode the features to the original resolution. To fully utilize both high- and low-level features, we follow the UNet-like structure~\cite{ronneberger2015u} to concatenate the feature from the encoder and previous layer of the decoder as the input of the next decoder layer. Each decoder layer is composed of three convolution layers with batch normalization. After getting the output of the decoder, a convolution layer (denoting as $\textit{Affinity}(\cdot)$) is employed to predict pixel-wise affinity kernels. It also predicts a refined flow to use in the subsequent propagation process by employing a separate convolution layer (denoting as $\textit{RefinedFlow}(\cdot)$), \ie,
\begin{equation}
    \begin{aligned}
        F_i^{\{c\}} &= \textit{Encoder}([I_i, \widehat{I}_{i-1}, M_i, \widehat{M}_{i-1}, \widetilde{M_i},\\ &\widetilde{{Op}}_{i-1 \to i},
            {Op}_{i-1 \to i}, G_i, \widehat{G}_{i-1}]), c \in \{1,2,3,4\},
    \end{aligned}
\end{equation}
\begin{equation}
        D_i^c = \textit{Decoder}([D_i^{c-1}, F_i^{4-c}]), c \in \{1,2,3,4\},
\end{equation}
\begin{equation}
\begin{aligned}
        \kappa_i &= \textit{Affinity}(D_i^4) \\
        B_i &= \textit{RefinedFlow}(D_i^4).
\end{aligned}
\end{equation}
The affinity matrix $\kappa_i$ is of size $[H_m, W_m, (2r+1)^2]$, where $r$ is the radius of affinity kernel, \ie, 4 in this paper. $B_i$ is the refined flow of size $[H_m, W_m, C]$, \ie, which also provides initial estimate for the out-of-boundary view. Then, we use the affinity matrix $\kappa_i$, the refined optical flow $B_i$, and the mask $\widetilde{M}_i$ to propagate the optical flow from pixels in $\widetilde{M}_i$ to the out-of-boundary pixels. Mathematically, this process can be formulated as follows:
\begin{equation}\label{eqn:CSPNPro}
\begin{aligned}
    B_{i}^{t+1}[u,v] = \hat{\kappa}_{i}[u,v,r] \cdot B_{i}^0[u, v] + \\
    \sum_{a,b=-r, a,b \neq 0}^{r} \hat{\kappa}_{i}[u,v,ar+b] \cdot B_{i}^{t}[u-a, v-b],
\end{aligned}
\end{equation}
\begin{equation}\label{eqn:CSPNPro2}
\begin{aligned}
    \hat{\kappa}_{i}[u,v,ar+b] &= \frac{\kappa_{i}[u,v,ar+b]}{\sum_{a,b \neq 0}\kappa_{i}[u,v,ar+b]},
\end{aligned}
\end{equation}
\begin{equation}\label{eqn:CSPNPro3}
\begin{aligned}
       \hat{\kappa}_{i}[u,v,r] &= 1 - \sum_{a,b \neq 0}\hat{\kappa}_{i}[u,v,ar+b],
\end{aligned}
\end{equation}
where $u$ and $v$ denote the 2D location of each pixel, $t$ denotes the number of iterations, and $B_{i}^0 \triangleq B_i$.

\subsection{Convergence Analysis}
We provide a brief theoretical analysis here to show that the propagation will not cause the propagated flow to explode as in ~\cite{liu2017learning,cheng2020learning}. For clarity, we denote $\lambda_{u,v}=\sum_{a,b \neq 0} \hat{\kappa}_{i}[u,v, ar+b]$, vectorize $B_i$ to the shape of $[H_m \times W_m, c]$, and use the same symbol without causing ambiguity.
Eq.~\eqref{eqn:CSPNPro} can be rewritten as:
\begin{equation}
\footnotesize
\begin{aligned}
    B^{t+1}_{i} &= \left[ \begin{array}{cccc}
0 & \hat{\kappa}_{i}[0, 0, 1 * r + 0] & \cdots & 0\\
\hat{\kappa}_{i}[1, 0, -1 *r+ 0] & 0 & \cdots & 0\\
\vdots & \vdots & \ddots & \vdots \\
\vdots & \vdots & \cdots & 0
\end{array} 
\right ] \cdot B^{t}_{i} \\
&+ \left[ \begin{array}{cccc}
1 - \lambda_{0,0} & 0 & \cdots & 0\\
0 & 1 - \lambda_{1, 0} & \cdots & 0\\
\vdots & \vdots & \ddots & \vdots \\
\vdots & \vdots & \cdots & 1 - \lambda_{H_m, W_m}
\end{array} 
\right ] \cdot B^{0}_{i} \\
&\triangleq AB_{i}^t+ (I-D)B_{i}^{0},
\end{aligned}
\label{eq:affinityPropagation}
\end{equation}
where $A$ and $D$ denote the former and latter matrices, respectively. The partial differential of Eq~\eqref{eq:affinityPropagation} w.r.t. $t$ is:
\begin{equation}
\label{eqn:GradH1}
    \partial_t B_{i}^{t+1} = A \partial_t B_{i}^t.
\end{equation}
Therefore, we have:
\begin{equation}
\small 
    || \frac{\partial_t B_{i}^{t+1}}{\partial_t B_{i}^t} || = ||A|| \leq \lambda_{max} = \textit{max}\sum_{a,b \neq 0}\hat{\kappa}_{i}[u,v,ar+b].
\end{equation}
Note that, the maximum gradient will always be less than 1 because we use the normalized kernel value $\hat{\kappa}$ during the propagation process. Thus the stability of the flow propagation based on affinity can be guaranteed. Besides, since the refined flow inside the shared view is much more reliable, we use a slow update strategy to preserve their flow values during the propagation process. To be specific, we use a ratio of $0.9$ for updating as follows:
\begin{equation}
    B_{i}^t = (1 - 0.9 \times \widetilde{M_i}) B_{i}^t + 0.9 \times B_{i}^0 \widetilde{M_i}.
\end{equation}
\subsection{Training Objective}
The loss function of the coarse alignment stage are the same as the DUT~\cite{xu2020dut}. For the fine alignment stage, we use robust L1 loss during training. Given the propagated optical flow $B_{i}^t$, we can extrapolate the pixel value outside the boundary in $\widehat{G}_{i-1}$, $\widehat{I}_{i-1}$, and $\widehat{M}_{i-1}$, respectively, \ie
\begin{equation}
    \widehat{G}_{i-1}^e = \textit{Extrapolate}(\widehat{G}_{i-1}, B_{i}^t).
\end{equation}
\begin{equation}
    \widehat{I}_{i-1}^e = \textit{Extrapolate}(\widehat{I}_{i-1}, B_{i}^t).
\end{equation}
\begin{equation}
    \widehat{M}_{i-1}^e = \textit{Extrapolate}(\widehat{M}_{i-1}, B_{i}^t).
\end{equation}
Then, we can calculate the L1 loss between the above predictions and their ground truth ($I_i^g$, $G_i^g$), \ie,
\begin{equation}
    L_I = |\widehat{I}_{i-1}^e - I_i^g + \epsilon| \odot  \hat{M}_{i-1}^e,
\end{equation}
\begin{equation}
    L_G = |\widehat{G}_{i-1}^e - G_i^g + \epsilon| \odot \hat{M}_{i-1}^e,
\end{equation}
where $\epsilon$ is a small value for stability, \ie, $10^{-12}$ in this paper. Note that there may be a trivial propagation solution, where very element in the extrapolated mask $\widehat{M}_{i-1}^e$ based on $B_{i}^t$ is zero. To address this issue, we add a regularization loss on $\widehat{M}_{i-1}^s$ as follows:
\begin{equation}
    L_M = \textit{MSE}(\widehat{M}_{i-1} \widehat{M}_{i-1}^e, \widehat{M}_{i-1}),
    \label{eq:mse_mask}
\end{equation}
which penalizes the shrinkage of the mask region after propagation. The final training objective function for the fine alignment stage is $L = L_I + 2 \times L_G + 2 \times L_M$, where the loss weights are set empirically.

Note that although we adopt supervised training for OVS, it does not require paired data. In our experiments, we prepare the training data by cropping the unstable frames. Specifically, we first randomly crop a region of size $800 \times 640$ from each unstable frame of size $1280 * 720$ and use it as the ground truth $I_i^g$. Then, we crop its central part of size $640 \times 480$ as $I_i$. The left surrounding part in $I_i^g$ is indeed the out-of-boundary view w.r.t. $I_i$.

\begin{figure*}[htbp] 
\centering 
    \includegraphics[width=0.94\linewidth]{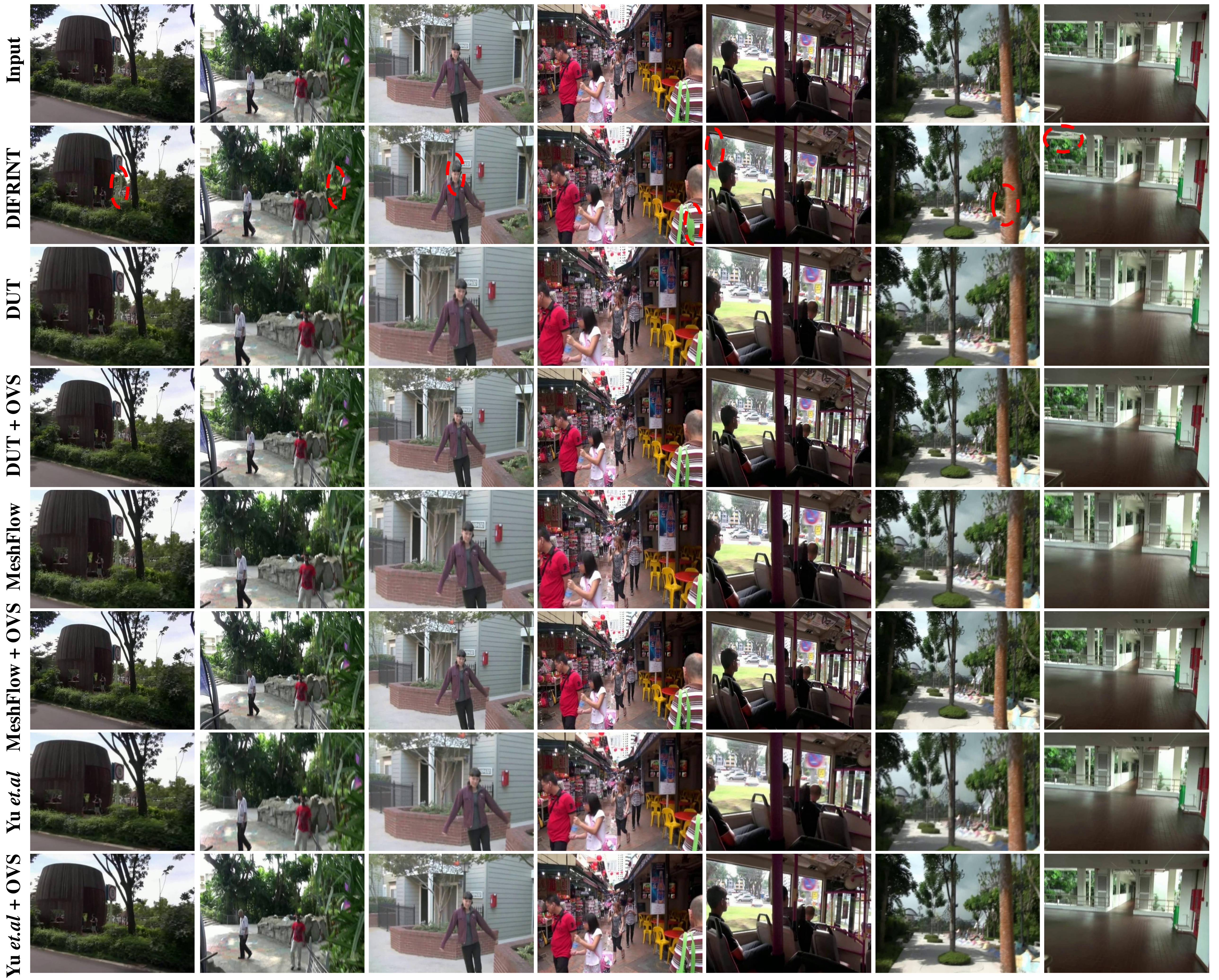}
    \caption{Subjective comparison of DIFRINT~\cite{Choi_TOG20}, DUT~\cite{xu2020dut}, MeshFlow~\cite{liu2016meshflow} and Yu \etal~\cite{yu2020learning}. ``+OVS'' denotes the results from corresponding stabilizer integrated with our OVS. Red circles highlight the ghost artifacts in DIFRINT's results.}
    \label{fig:SubCompare1}
\end{figure*}

\section{Experiments}

\subsection{Implementation Details}

We only use the unstable videos in the DeepStab~\cite{wang2018deep} dataset for training and validation. Specifically, fifty videos are used for training and other ten videos are used for validation. We use the Adam~\cite{kingma2015adam} optimizer with beta (0.9, 0.99) during training. The learning rate is set to 2e-4 and decays by 0.5 every 30 epochs. We train the coarse alignment stage for 50 epochs and the fine alignment stage for 200 epochs, respectively. The training process takes about 36 hours on a single NVIDIA V100 GPU. We test the performance of our OVS model on the NUS~\cite{liu2013bundled} dataset.

\subsection{Quantitative Evaluation}

\begin{table}[htbp]
  \centering
  \caption{Average metrics on the NUS~\cite{liu2013bundled} dataset. * means the DUT stabilizer integrated with our full-frame version OVS.}
    \setlength{\tabcolsep}{0.012\linewidth}{\begin{tabular}{l|rrr}
    \hline
          &\multicolumn{1}{c}{\textbf{Cropping}} & \multicolumn{1}{c}{\textbf{Distortion}} & \multicolumn{1}{c}{\textbf{Stability}} \\
    \hline
    DIFRINT & 0.949  & 0.854  & 0.823  \\
    \hline
    PWStabNet & 0.877  & 0.924  & 0.830 \\
    PWStabNet + \textbf{OVS} & 0.959  & \textbf{0.957}  & 0.832 \\
    \hline
    Yu \etal & 0.827  & 0.722  & 0.814 \\
    Yu \etal + \textbf{OVS} & 0.922  & 0.784  & 0.834 \\
    \hline
    StabNet &  0.676  & 0.731  & 0.741  \\
    StabNet + \textbf{OVS} & 0.763  & 0.829  & 0.748  \\
    \hline
    MeshFlow &  0.770  & 0.673  & 0.813  \\
    MeshFlow + \textbf{OVS}  & 0.898  & 0.683  & 0.823  \\
    \hline
    DUT  & 0.867  & 0.895  & 0.845 \\
    DUT + \textbf{OVS} & \underline{0.967}  & 0.926 & \underline{0.847} \\
    DUT + \textbf{OVS*} & \textbf{0.999}  & \underline{0.944} & \textbf{0.849} \\
    \hline
    \end{tabular}}%
  \label{tab:ObjMetrics}%
\end{table}%

To demonstrate the versatility and effectiveness of our proposed OVS, we select several representative warping-based stabilizers and integrated OVS into these stabilizers as a plug-and-play module. Specifically, PWStabNet~\cite{zhao2020pwstablenet} and Yu \etal~\cite{yu2020learning} utilize pixel-based waring for stabilization while StabNet~\cite{wang2018deep}, Meshflow~\cite{liu2016meshflow}, and DUT~\cite{xu2020dut} are grid-based warping stabilizers. OVS is integrated with these stabilizers by replacing the warping steps and keeping the rest of each method unchanged.

The results are summarized in Table~\ref{tab:ObjMetrics}. We use three metrics following Bundled~\cite{liu2013bundled} for performance evaluation, including cropping ratio, distortion, and stability, which are better for larger values. DIFRINT is the interpolation-based method, which performs better in terms of cropping ratio while falls behind warping-based methods like PWStabNet and DUT in terms of stability and distortion. It is noteworthy that the cropping ratio of DIFRINT is smaller than 1 although it does not require cropping. We suspect that this is because the stabilizer learned a small zoom in effect during the iterative interpolation process, as shown in the Figure~\ref{fig:SubCompare1}. More analysis is provided in the \supp{supplementary material}. 

In addition, it can be seen that OVS can significantly improve the cropping ratio of all the warping-based methods. Meanwhile, the distortion and stability of these methods also increase as a by-product since larger cropping ratio means less cropping and resizing of the stabilized frames, therefore reducing the jitter amplification effect. Beside, since cropping aims to remove the black holes around the boundary, it does not guarantee to keep the original frame aspect ratio. Consequently, when the holes dominate in one direction, there may be large distortions in the result, \ie, a lower distortion metric. In contrast, after extrapolating the out-of-boundary view via the proposed OVS, the frame aspect ratio can be better preserved, resulting in larger distortion metrics. We also notice that when applying our OVS in DUT, the cropping ratio is 0.967, which is close to 1. Most of the holes lie in texture-less regions like sky. We further use a simple trick to fill in the holes using nearest neighbor interpolation during warping. This trick leads to full-frame stabilization results as shown in the last row in Table~\ref{tab:ObjMetrics}.

\subsection{Qualitative Evaluation}

We also provide some subjective results of DUT~\cite{xu2020dut}, Meshflow~\cite{liu2016meshflow}, and Yu \etal~\cite{yu2020learning} with and without our OVS as well as DIFRINT~\cite{Choi_TOG20} for comparison in Figure~\ref{fig:SubCompare1}. It can be seen that without the OVS, there is a significant content losses in the DUT, Meshflow, and Yu \etal's results due to the absence of out-of-boundary views during warping. This content loss can be mitigated by extrapolating the out-of-boundary view using our OVS as shown in the fourth, sixth, and last rows. The results of DIFRINT suffer from large distortions, as highlighted by the red circle, especially when there are severe jitters between neighboring frames and dynamic objects. It is because the optical flow extracted for interpolation is not accurate due to large jitters and dynamic objects, leading to ghost artifacts around object boundaries. Besides, the results in DIFRINT suffer from small content loss, \eg, the bottom left flag in the second column and the zoom-in effect in the sixth column. In contrast, OVS mitigates these issues, as shown in the results of DUT+OVS. Based on both the quantitative and qualitative evaluation results, OVS demonstrates its effectiveness in helping warping-based stabilizers to better video stabilization results with a high cropping ratio, \eg, DUT+OVS achieves near full-frame stabilization results.

\subsection{Ablation Study}

\subsubsection{Influence of Coarse and Fine Alignment}

\begin{figure}[htbp] 
\centering 
    \includegraphics[width=0.95\linewidth]{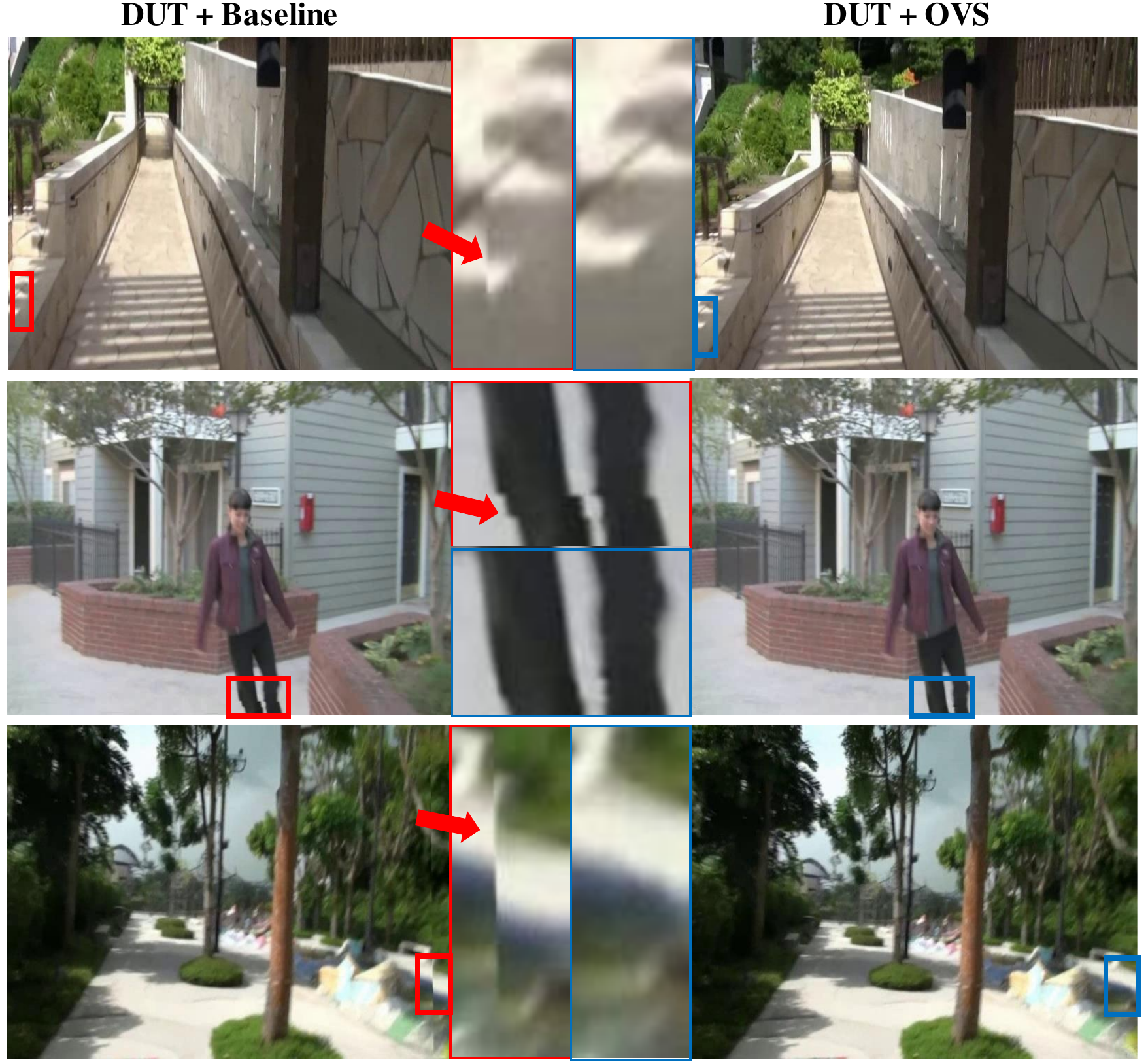}
    \caption{Results of DUT with a SOTA image alignment method \cite{zhang2020content} and our OVS. Red arrows indicate the misalignment.}
    \label{fig:BaseCompare2}
\end{figure}

\begin{table}[htbp]
  \centering
  \caption{Ablation study of OVS.}
    \setlength{\tabcolsep}{0.015\linewidth}{\begin{tabular}{l|rrrrr}
    \hline
          & \multicolumn{1}{l}{PSNR} & \multicolumn{1}{l}{SSIM} & \multicolumn{1}{l}{Crop} & \multicolumn{1}{l}{Distortion} & \multicolumn{1}{l}{Stability} \\
    \hline
    Baseline & 15.71 & 0.74 & 0.961 & 0.904 & 0.845 \\
    \hline
    CoarseOnly & 17.66  &  0.61  & 0.961  & 0.917  & 0.848  \\
    FineOnly & 18.60  & 0.72  &  0.868 & 0.884 & 0.845 \\
    OVS  & 22.84 & 0.82  & 0.967  & 0.926  & 0.847  \\
    \hline
    \end{tabular}}%
  \label{tab:CRAblation}%
\end{table}%
\begin{figure}[htbp] 
\centering 
    \includegraphics[width=0.96\linewidth]{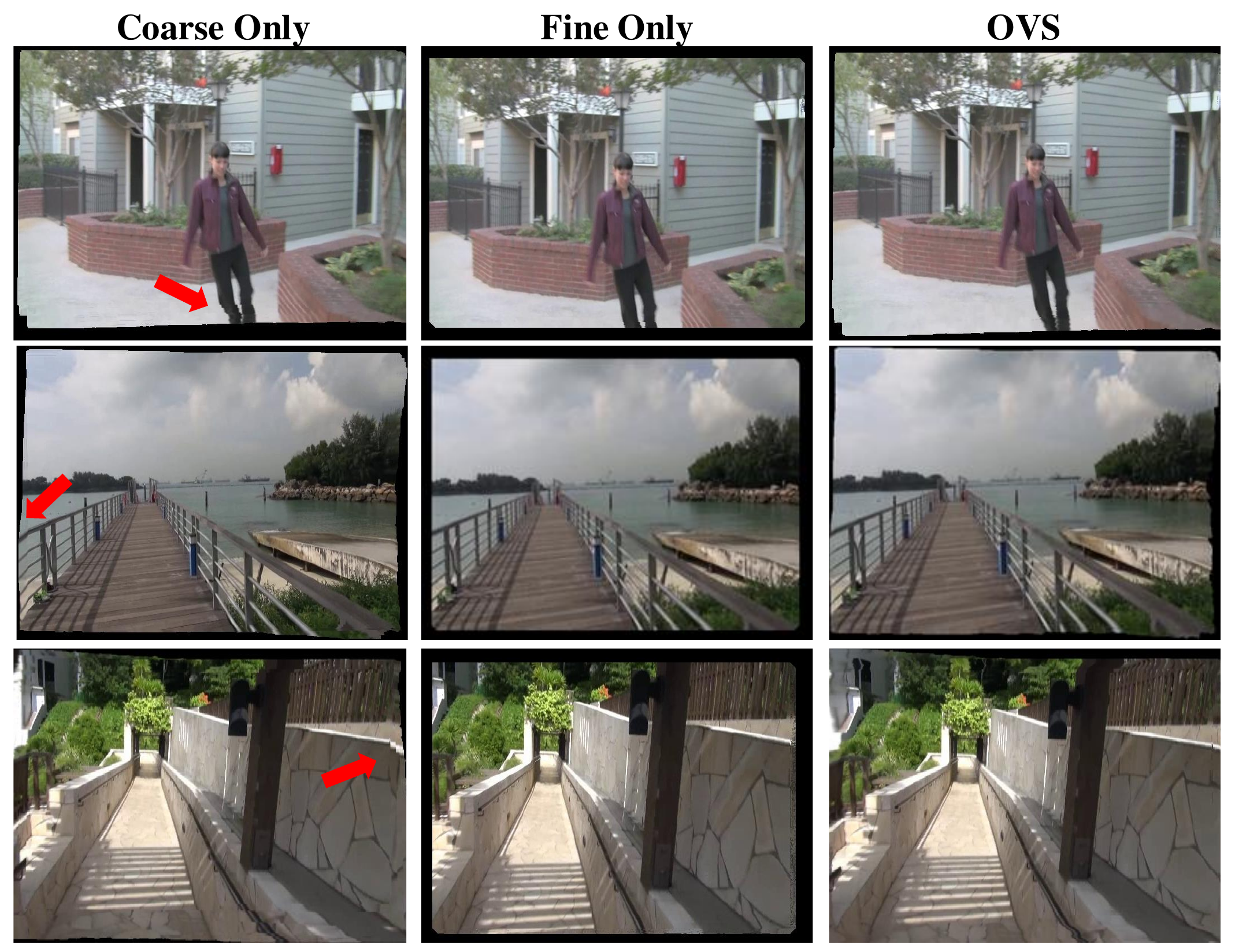}
    \caption{Visual comparison of coarse alignment, fine alignment, and OVS. Red arrows indicate the misalignment.}
    \label{fig:RoughRefine}
\end{figure}

We select the SOTA image alignment method in \cite{zhang2020content} as a comparison to our OVS, which is a deep learning based method and specifically designed for adjacent frames alignment. In addition, we isolate the coarse alignment stage and the fine alignment stage from our OVS to compare their results with the complete version of OVS. Specifically, we choose DUT as the basic stabilizer and use the aligned frames from different models for warping. Their PSNR and SSIM metrics on the validation set are summarized in Table~\ref{tab:CRAblation}. It can be seen that 1) the baseline alignment method \cite{zhang2020content} can achieve good alignment results in terms of the SSIM scores and help DUT to improve the cropping ratio, demonstrating its effectiveness for image alignment. Our coarse alignment module can also help DUT improve the cropping ratio and distortion metric as well as obtains a better PSNR score than \cite{zhang2020content}. However, its SSIM score is the worst among all the models, implying that there are some structural misalignment in the results. Our fine alignment module can obtain better out-of-boundary view synthesis results than the coarse alignment module in terms of both PSNR and SSIM metrics. However, it does not help DUT to achieve better stabilization results, implying that it can only serve as a refinement module. After combining both modules together, our OVS achieves the best view synthesis performance among all the models as well as significantly improve the performance of DUT for video stabilization. 

Some visual results are shown in Figure~\ref{fig:BaseCompare2} and Figure~\ref{fig:RoughRefine}. The baseline method \cite{zhang2020content} produces significant discontinuities around the boundary of sharp edges and dynamic objects in its stabilized results, while our OVS has no such distortions. Similar discontinuities can be observed in the results of only using coarse alignment for warping, implying the necessity of fine alignment. However, only using fine alignment leads to less out-of-boundary view as shown by the large holes around the boundaries. These results confirm the complementarity between coarse alignment and fine alignment in our OVS, which help synthesize large out-of-boundary views with less distortions collaboratively.

\subsubsection{Analysis of the Influence of Iterations}

\begin{table}[htbp]
  \centering
  \caption{The influence of different iterations.}
    \setlength{\tabcolsep}{0.030\linewidth}{\begin{tabular}{r|rrr}
    \hline
    Iteration  & \multicolumn{1}{l}{Crop} & \multicolumn{1}{l}{Distortion} & \multicolumn{1}{l}{Stability} \\
    \hline
    0 & 0.867 & 0.895 & 0.845 \\
    \hline
    5 & 0.938 & 0.930 & 0.847  \\
    10 & 0.967 & 0.926 & 0.847 \\
    15 & 0.970 & 0.925 & 0.847 \\
    \hline
    \end{tabular}}%
  \label{tab:IterAblation}%
\end{table}%

We investigate the influence of iterations when employing OVS in warping-based stabilizers, \eg, taking DUT as an exemplar stabilizer. The stability, distortion, and cropping ratio at different settings are shown in Table~\ref{tab:IterAblation}. It can be seen that as the number of iterations increases, the cropping ratio first increases and then almost reaches saturation. The distortion metric increases significantly at the beginning and then slightly decreases while the stability metric keep almost the same. As the number of iterations increases, more distant frames are aligned to the current reference frame for out-of-boundary view synthesis, leading to a larger area. Besides, there may be subtle misalignment accumulated during the iterations, especially misalignment from distant frames. Nevertheless, distant frames indeed contribute less to synthesis since they have less shared areas with the reference frame. Therefore, the distortion metric is slightly decreased as the number of iterations increases.

\section{Limitation and discussion}
Due to variations in adjacent frames, such as illumination, dynamic objects, and noise, some visible seams may exist in the stabilized frames, degrading the visual experience. Such discontinuities can be refined using an encoder-decoder network like \cite{lempitsky2018deep}. It is noteworthy that we use the widely used PWCNet \cite{sun2018pwc} for optical flow estimation following \cite{Choi_TOG20,xu2020dut} and an simple encoder-decoder network for fine alignment and affinity propagation, which can not be claimed as a major contribution. Actually, the major contrition of this work is that we provide a fresh perspective for warping-based methods towards full-frame video stabilization, \ie, explicitly extrapolating the requisite out-of-boundary view during warping. In addition, we exploit the spatial coherence in the video to achieve this goal via a simple coarse-to-fine scheme. In the future, we plan to devise a more effective end-to-end model for better out-of-boundary view synthesis and video stabilization.

\section{Conclusion}

This paper presents a new Out-of-boundary View Synthesis (OVS) method that can help warping-based stabilizers to achieve near full frame stabilization with less distortions and better stability. OVS exploits the spatial coherence in the video to effectively align adjacent frames to reference frames and synthesize the out-of-boundary view, therefore benefiting the warping process in warping-based stabilizers directly by providing requisite candidate pixels. It can serve as a plug-and-play module and significantly improve both pixel- and grid-based warping stabilizers. We hope this study can provide valuable insights to the community and inspire follow-up research from a different but promising direction for video stabilization.

\noindent\textbf{Acknowledgement} Mr. Yufei Xu and Dr. Jing Zhang are supported by the ARC project FL-170100117.

\clearpage

{\small
\bibliographystyle{ieee_fullname}
\bibliography{egbib}
}

\end{document}